\documentclass[letterpaper, 10 pt, conference]{URL-ieeeconf}
\IEEEoverridecommandlockouts    
\usepackage{graphics}           
\usepackage{times}              
\usepackage{amsmath}            
\usepackage{amssymb}            
\usepackage{graphicx}
\usepackage{algorithm}
\usepackage[noend]{algpseudocode}
\usepackage{booktabs}
\usepackage{color}
\usepackage{multirow}
\usepackage{subcaption}
\usepackage{rotating}
\usepackage{cite}
\definecolor{instructioncolor}{rgb}{.5,.5,.5}

\def\eqref#1{(\ref{#1})}

\newcommand{\rom}[1]{\uppercase\expandafter{\romannumeral #1\relax}}

\makeatletter
\usepackage{xspace}
\DeclareRobustCommand\onedot{\futurelet\@let@token\@onedot}
\def\@onedot{\ifx\@let@token.\else.\null\fi\xspace}

\makeatother

\usepackage{array}
\newcolumntype{L}[1]{>{\raggedright\let\newline\\\arraybackslash\hspace{0pt}}m{#1}}
\newcolumntype{C}[1]{>{\centering\let\newline\\\arraybackslash\hspace{0pt}}m{#1}}
\newcolumntype{R}[1]{>{\raggedleft\let\newline\\\arraybackslash\hspace{0pt}}m{#1}}

\usepackage{cite}
\usepackage[hidelinks]{hyperref}

\title{\LARGE \bf OpenHEART\@: Opening Heterogeneous Articulated Objects \\with a Legged Manipulator}
 
\author{Seonghyeon Lim$^1$, Hyeonwoo Lee$^1$, Seunghyun Lee$^1$, \\ I Made Aswin Nahrendra$^{2 \dagger}$, and Hyun Myung$^{1 *}$
  \thanks{$^*$Corresponding author: Hyun Myung}
  \thanks{$\dagger$Work done during the time at 1.}
  \thanks{
    $^1$Seonghyeon Lim, Hyeonwoo Lee, Seunghyun Lee, and Hyun Myung are with the School of Electrical Engineering, KAIST (Korea
    Advanced Institute of Science and Technology), Daejeon, 34141, Republic of Korea.
    {\tt\scriptsize \{shlim, hyeonwoolee, kevin9709, hmyung\}@kaist.ac.kr} \hfill \break
    \indent $^2$I Made Aswin Nahrendra is with KRAFTON, Seoul, 06142, Republic of Korea. 
    {\tt\scriptsize anahrendra@krafton.com} \hfill \break
  }
}

\begin{document}
\maketitle
\thispagestyle{empty}
\pagestyle{empty}

\newtheorem{remark}{Remark}[section]

\begin{abstract}
Legged manipulators offer high mobility and versatile manipulation. 
However, robust interaction with heterogeneous articulated objects, such as doors, drawers, and cabinets, remains challenging because of the diverse articulation types of the objects and the complex dynamics of the legged robot.
Existing reinforcement learning (RL)-based approaches often rely on high-dimensional sensory inputs, leading to sample inefficiency. 
In this paper, we propose a robust and sample-efficient framework for opening heterogeneous articulated objects with a legged manipulator. 
In particular, we propose Sampling-based Abstracted Feature Extraction (SAFE), which encodes handle and panel geometry into a compact low-dimensional representation, improving cross-domain generalization.
Additionally, Articulation Information Estimator (ArtIEst) is introduced to adaptively mix proprioception with exteroception to estimate opening direction and range of motion for each object. 
The proposed framework was deployed to manipulate various heterogeneous articulated objects in simulation and real-world robot systems.
Videos can be found on the project website: \url{https://openheart-icra.github.io/OpenHEART/}.
\end{abstract}

\newcommand{\extest}{\hat{\boldsymbol{\alpha}}^\text{ext}_t}
\newcommand{\propest}{\hat{\boldsymbol{\alpha}}^\text{prop}_t}
\newcommand{\mixest}{\hat{\boldsymbol{\alpha}}^\text{mix}_t}
\newcommand{\artinfo}{\boldsymbol{\alpha}_t}
\newcommand{\mixratio}{\boldsymbol{\gamma}_t}

\newcommand{\command}{\textbf{c}_t}
\newcommand{\lastcommand}{\textbf{c}_{t-1}}
\newcommand{\lastlastcommand}{\textbf{c}_{t-2}}

\newcommand{\commandee}{\textbf{c}^\text{EE}_t}
\newcommand{\lastcommandee}{\textbf{c}_{t-1}}

\newcommand{\shapingcoef}{\lambda_t}
\newcommand{\generatio}{test/train}

\newcommand{\tbu}[1]{\textcolor{cyan}{#1}}

\section{Introduction}
\label{sec:introduction}
A legged manipulator, combining the locomotion capability of a quadruped robot with a manipulating arm, offers the versatility needed for a wide range of tasks in everyday environments~\cite{liu2025mlm,ha2025umi,bruedigam2025jacta,liu2025visual}. 
The everyday tasks usually involve opening heterogeneous articulated objects, such as doors, cabinets, and drawers.
The heterogeneous articulated objects can vary in appearance and joint directions.
However, opening heterogeneous articulated objects with a legged manipulator remains challenging.
Not only the different handle shapes and object articulation types require different manipulation strategies, but also the robot itself has complex dynamics due to its high degree of freedom (DoF) and floating base configuration.

Recently, reinforcement learning (RL)-based methods have demonstrated remarkable success in contact-rich tasks~\cite{cheng2024extreme, lee2020learning, nahrendra2023dreamwaq, ma2023learning, lim2024learning}, including articulated object manipulation~\cite{zhang2025learning, sleiman2025guided, li2024unidoormanip, geng2023partmanip, urakami2019doorgym}.
However, legged manipulators are usually limited to homogeneous door types~\cite{zhang2025learning, sleiman2025guided}.
Because these methods mainly focused on a single type of door with a similar shape, the position of the handle and doorway is sufficient to represent objects~\cite{zhang2025learning}.
Unlike homogeneous objects, heterogeneous articulated objects can be represented with other geometric features, such as the shape of the handle and panel. 
These representations are essential for the robot to generate appropriate motions to manipulate the objects.
To this end, high-dimensional exteroception such as point clouds~\cite{li2024unidoormanip,geng2023partmanip} and images~\cite{xiong2024adaptive} has often been used to represent the object.
Although such high-dimensional inputs provide abundant information about the object, they are inefficient for learning contact-rich tasks with legged manipulators.
The complexity of the robot's dynamics and task could increase the number of required training samples, making sample efficiency more important.
However, training an RL policy directly from high-dimensional observations is typically sample inefficient~\cite{banino2021coberl, lake2017building}.

\begin{figure}[t] 
    \centering
    \hspace{-8mm}
    \captionsetup{font=small}
    \includegraphics[width=0.44\textwidth]{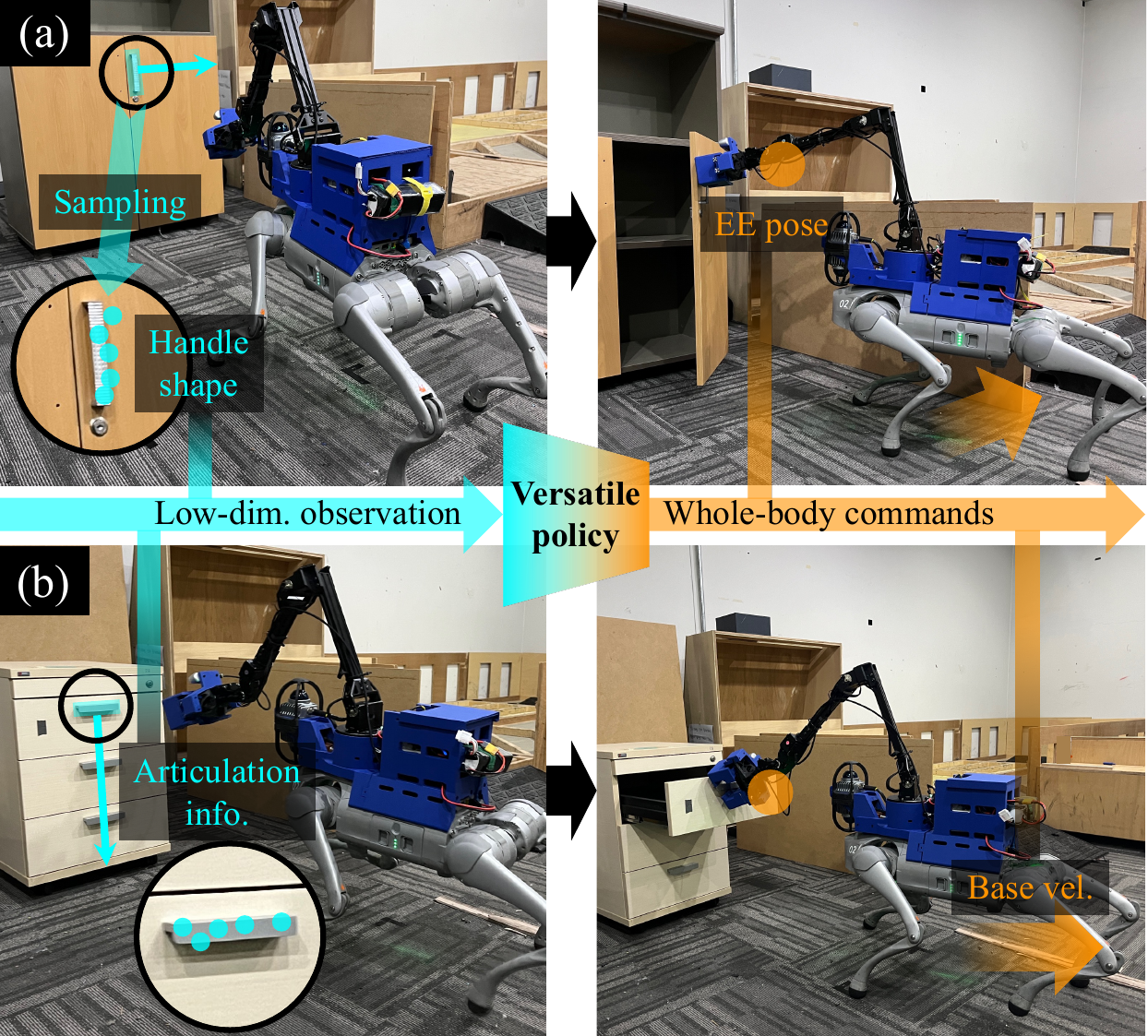}
    \caption{
 We present a framework for opening heterogeneous articulated objects with a legged manipulator.
 The robot can open diverse articulated objects, such as (a) a revolute cabinet with a vertical handle and (b) a prismatic drawer with a horizontal handle without any object-specific models.
 }
    \label{fig:title}
    \vspace{-11mm}
\end{figure}

\begin{figure*}[t] 
\vspace{2mm}
\centering
\hspace*{-10mm}
    \includegraphics[width=0.81\textwidth]{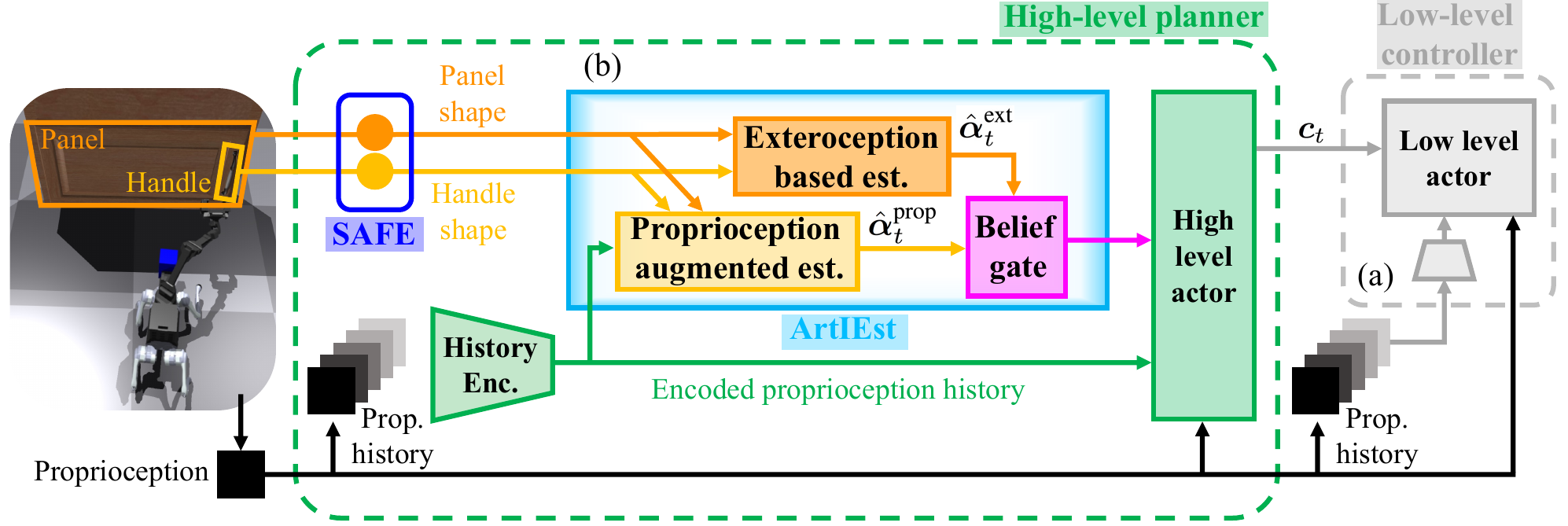}
    \vspace{-1mm}
    \captionsetup{font=small}
    \caption{
The framework is hierarchically structured, with a high-level planner and (a) a low-level controller.
SAFE is introduced to efficiently represent the object shape, while mitigating overfitting.
(b) ArtIEst estimates the articulation information by adaptively mixing exteroception and proprioception.
The history encoder extracts features from proprioception history.
 }
    \label{fig:heart_framework}
    \vspace{-10.5 mm}
\end{figure*}

In this study, we propose a framework for opening heterogeneous articulated objects using a legged manipulator.
We introduce a low-dimensional representation of objects that contains the geometric features, such as handle shape and articulation information, to open heterogeneous articulated objects with a single versatile policy (Fig.~\ref{fig:title}).
The articulation information includes the joint direction of the object and the distance of the handle from the joint axis.
Such features are required to estimate the grasping pose as well as the direction and range of opening motion.
We validate that the proposed low-dimensional representation enables learning a versatile policy to open various heterogeneous articulated objects with a legged manipulator.

Our contributions can be summarized as follows:
\begin{itemize}
\item A hierarchical framework is proposed to open heterogeneous articulated objects with a legged manipulator.
To the best of our knowledge, this is the first approach that achieves autonomous manipulation of heterogeneous articulated objects with a legged manipulator without a precise object model.
\item Articulation Information Estimator (ArtIEst) is proposed to attain object features required to infer the opening motion.
ArtIEst adaptively fuses proprioception with exteroception, achieving a lower estimation error than using exteroception only and an estimator that monolithically fuses both sensor modalities.
\item Sampling-based Feature Extraction (SAFE) is used to abstract object shapes into low-dimensional representations and reduce overfitting through a simple post-processing step.

\end{itemize}

\section{Related Work}
\label{sec:related_work}
\textbf{RL-based Articulated Object Manipulation.} \quad
RL has been widely applied to articulated object manipulation across various robotic platforms~\cite{geng2023partmanip, gu2017deep, sharma2021learning, li2024unidoormanip, rajeswaran2018learning, urakami2019doorgym, xiong2024adaptive, schwarke2023curiosity, luo2024human, ma2024hierarchical}.
However, most of these works focus on fixed-base manipulators~\cite{ma2024hierarchical, dalal2024local, sharma2021learning, gu2017deep, geng2023partmanip, li2024unidoormanip} or wheeled mobile manipulators~\cite{xiong2024adaptive}, which are not subject to floating-base configurations and generally have lower DoF than legged manipulators.
In contrast, legged manipulators involve floating-base configurations and multi-contact dynamics, which makes learning opening tasks more challenging.
These complex dynamics of robots can lead to a larger number of required training samples; therefore, sample efficiency can be crucial.
Zhang~\textit{et~al}.~\cite{zhang2025learning} and Sleiman~\textit{et~al}.~\cite{sleiman2025guided} demonstrate that legged manipulators can open a single type of door with various dimensions.
Zhang~\textit{et~al}.~\cite{zhang2025learning} represent the door using the position of the handle and doorway.
Although this representation is sample efficient and sufficient for opening a homogeneous door, it is insufficient for heterogeneous articulated objects with various shapes and articulation types.
In contrast, we propose a low-dimensional representation that can efficiently encode relevant features for opening heterogeneous articulated objects.

\textbf{Object Articulation Estimation.} \quad
Exteroception-based methods~\cite{wang2024rpmart, yu2024gamma, zeng2021visual, eisner2022flowbot3d, li2020category} can estimate articulation information without physical interaction.
However, such methods can suffer from visual ambiguity in estimation when the appearance of an object implies multiple possible articulation models, especially with symmetric objects.
In contrast, proprioception-based methods~\cite{lips2023revisiting, zhang2025learning} are proposed to estimate the articulation information after grasping.
However, such approaches are applicable only when the robot is in contact with the object.
Additionally, they do not leverage observable geometric cues.
Buchanan~\textit{et~al}.~\cite{buchanan2024online} address these limitations by fusing both sensor modalities using a factor graph framework.
However, applying such a model-based approach to a legged manipulator is challenging.
The high degrees of freedom, floating-base dynamics, and contact-rich interactions complicate accurate model-based estimation.
Zhang~\textit{et~al}.~\cite{zhang2025learning} present a learning-based approach combining both inputs, enabling a legged manipulator to estimate door types, such as push/pull and left/right.
However, their approach is limited to a single articulation type and monolithically fuses both sensor modalities.
In comparison, our proposed method, ArtIEst, can estimate various articulation types by adaptively fusing exteroception and proprioception based on the contact state.

\section{Methodology}
\label{sec:method}
Our framework is hierarchically structured with a low-level controller and a high-level planner. 
The low-level controller is responsible for generating joint target positions of the legged manipulator to track the given command, $\command$.
The command consists of the end-effector (EE) pose, the velocity of the quadruped base, and a boolean for gripper closing.
The high-level planner generates an appropriate $\command$ to open the target articulated object.
We pre-trained the low-level controller (Fig.~\ref{fig:heart_framework} (a)) with RL, using a proprioception history encoder to estimate environmental states~\cite{nahrendra2023dreamwaq, fu2023deep}.
Subsequently, the high-level planner was trained with heterogeneous articulated objects.

\subsection{SAFE: Sampling-based Abstracted Feature Extraction}

SAFE is designed to represent the object shape with a low-dimensional feature, while reducing the risk of overfitting to the training object sets.
For the opening task, one important feature is the relative length of each side of the handle and panel.
In particular, the relative length of the handle determines the grasping strategy, and that of the panel implies the opening direction.
Thus, we abstracted the detailed appearance of the handle and panel into an enveloping cuboid, while preserving relative length features.
During the training, such cuboids are provided from part annotations in the dataset~\cite{geng2023gapartnet,geng2023partmanip}.
Additionally, such an enveloping cuboid can be readily obtained during deployment, using off-the-shelf object detection algorithms~\cite{redmon2016you, wang2025yoloe}.


One major challenge in transferring an RL agent to the real world is the distributional difference between training and test objects.
Although numerous articulated object datasets~\cite{mu2021maniskill, liu2022akb} can be simulated during training, the diversity of assets is substantially lower than that of real-world objects.
Therefore, if an RL agent overfits to the training assets, it may not generalize well to diverse real-world objects.

The risk of overfitting can be mitigated when the distributional difference between training and test sets is reduced.
We selected Kullback-Leibler (KL) divergence as the metric to measure this discrepancy.

\begin{figure}[t!]
\vspace{2mm}
    \centering
    \hspace{-8mm}
    \captionsetup{font=small}
    \includegraphics[width=0.39\textwidth]{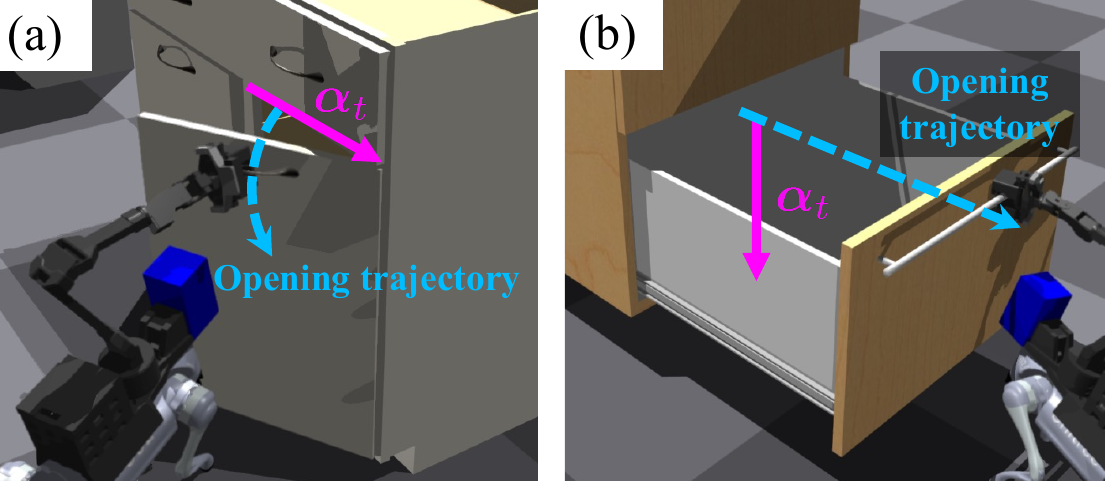}
    \vspace{-1mm}
    \caption{
Defined articulation information (pink solid arrow) for objects with (a) a revolute joint and (b) a prismatic joint.
The expected opening trajectory is shown as a blue dashed arrow.
 }
    \label{fig:artinfo}
    \vspace{-11mm}
\end{figure}

\begin{remark}
Let $P_C$ and $Q_C$ be the distributions of cuboid corner points $C \in \mathbb{R}^{8\times 3}$ corresponding to the handle shapes in $P$ and $Q$, respectively.
The data processing inequality~\cite{cover1991elements} states that no processing can increase the distinguishability of random variables.
As a result, any post-processing either decreases or preserves the KL divergence between distributions.
Let $f_C: \mathbb{R}^{8\times 3} \rightarrow \mathbb{R}^{N\times 3}$ be any post-processing function, then the following holds:
\begin{equation}
D_{KL}(P_C \| Q_C) \ge D_{KL}(f_C(P_C) \| f_C(Q_C)).
\end{equation}
Equality holds if $f_C$ preserves all the information distinguishing $P_C$ and $Q_C$.
\end{remark}

In the post-processing step, we replaced the corner points of each cuboid with points randomly sampled from its interior according to a uniform distribution.
This post-processing does not preserve all the information about the distinction between $P_C$ and $Q_C$.
Therefore, the KL-divergence between the transformed distributions is expected to be smaller than that between the original distributions.
In this way, the distributional difference between the training and test sets can be reduced.

Additionally, the ordering of points was randomly shuffled during the sampling process.
This randomness makes it difficult for the agent to learn a consistent mapping from the observation to the action.
Therefore, considering that three axes can always be defined to align with the cuboid's sides, we sorted the sampled points along the lateral axis of the cuboid.
In this way, the sampled points can preserve the order consistency.

To handle the motion blur in the image and the occlusion during manipulation, we captured the enveloping cuboids and sampled the points at the beginning of each episode, and used them throughout the episode.
All sampled points were transformed to the robot's base frame.

\subsection{ArtIEst: Articulation Information Estimator}

We defined articulation information, $\artinfo$, which can provide appropriate motion instructions after grasping.
In the case of revolute joints, $\artinfo$ is defined as the perpendicular vector from the center of the handle to the joint axis of the object.
Therefore, the direction of the vector implies the opening direction, and the scale indicates the range of motion.
For prismatic joints, we use a downward unit vector, which has the same direction as $\artinfo$ for the revolute-down joint, since their motion during the opening is similar.
Fig.~\ref{fig:artinfo} illustrates the defined $\artinfo$ for revolute and prismatic joints.

ArtIEst is proposed to estimate $\artinfo$ without a precise model of the object.
ArtIEst consists of an exteroception-based estimator, a proprioception-augmented estimator, and a belief gate that adaptively mixes the two estimations.
The overall structure of ArtIEst is illustrated in Fig.~\ref{fig:heart_framework} (b).
All modules in ArtIEst were concurrently trained with other components of the high-level planner, and the mean squared error (MSE) between the estimated and ground truth (GT) $\artinfo$ was used as the training loss.

\subsubsection{Exteroception-based Estimator}
The exteroception-based estimator was trained to estimate $\extest$~from geometric characteristics of the object.
The relative position of the handle on the panel can be an important cue for $\extest$.
For example, when a handle is placed on the left side of the panel, $\artinfo$~is typically determined as a vector pointing to the right side of the panel.
In contrast, when the same handle is placed on the right side of the panel, $\artinfo$~should be defined as a vector pointing to the left side.
Thus, we segment the panel and handle, and apply SAFE to each part independently.
It differs from previous approaches~\cite{yu2024gamma, wang2024rpmart} that estimate object joint information from the holistic point cloud of the object.
Subsequently, SAFE features from both parts are concatenated and fed into the estimator.
Additionally, we discovered that including the robot's orientation and EE pose in the input improves the estimation accuracy, thus they are also included in the input.

\subsubsection{Proprioception-augmented Estimator}
Although the exteroception-based estimator can infer $\artinfo$ before manipulation, estimation can be ambiguous when visual features suggest multiple candidates.
For instance, a cabinet with a horizontally elongated handle at the upper center may appear to open left, right, or downward.
To resolve such visual ambiguities, the proprioception-augmented estimator incorporates proprioceptive information with exteroceptive information during manipulation.
The estimator takes the current proprioception, encoded proprioception history, and SAFE features of the handle and panel as input.

However, proprioception can provide meaningful information about $\artinfo$ only when there is contact between the robot and the object.
Additionally, we find that using the errors of the proprioception-augmented estimation, $\propest$, as training signals without contact can lead to incorrect predictions.
Thus, we trained the proprioception-augmented estimator only while the door was opening.
Specifically, when the door was closed, the MSE loss of $\propest$ was not used for training.

\subsubsection{Belief Gating Mechanism}
Exteroception-based estimation is appropriate when there is no interaction with the object, whereas proprioception-augmented estimation could be more accurate while the robot is opening the object.
Thus, we implemented a belief gating mechanism to adaptively mix the two estimations.
The belief gate predicts the linear interpolation ratio, $\mixratio$, between the two estimations.
Subsequently, the mixed estimation is calculated as $\mixest = (1-\mixratio)\extest + \mixratio\propest$, where $0 \leq \mixratio \leq 1$.
The belief gate takes the encoded proprioception history and both estimations as input.

\subsection{High-level Actor}
\label{sec:high_level_policy}

The high-level actor was trained to infer appropriate commands for opening the target articulated object.
A history encoder encodes the proprioception history for the actor.
The encoder was trained with the $\beta$-variational autoencoder (VAE) loss~\cite{kingma2013auto, higgins2017beta}, while the corresponding decoder reconstructed the last proprioception.

\subsubsection{High-level Actor Observation}
\newcommand{\normmove}{\xi^\text{norm}_t}
\newcommand{\movement}{\xi_t}

\newcommand{\EEpos}{\textbf{p}^\text{EE}_t}
\newcommand{\handlepos}{\textbf{p}^\text{H}_t}
\newcommand{\EEori}{\theta_t}



\begin{table}[t!]
    \vspace{3mm}
    \centering
    \captionsetup{font=small}
    \caption{Reward function elements.
    $\EEpos$ and $\handlepos$ denote the position of the end-effector and handle, respectively.
    $\EEori$ is the angle between the dorsal-palmar axis of EE and the longer side of the handle.
    $\mathbb{I}\{\cdot\}$ is the indicator function, which returns one if the condition is satisfied, and zero otherwise.
 }
 \vspace{-1mm}
 {\scriptsize
    \begin{tabular}{llc}
        \toprule
        Reward & Equation & Weight \\
        \midrule
        Opening & $\normmove$ & 6.5 \\
        \midrule
        EE approach & 
        $\begin{cases}
        \text{exp}(-4\|\EEpos - \handlepos\|^2 ), & \text{if } \normmove < 0.2 \\
        1, & \text{otherwise}
        \end{cases} $ 
        & 1.9 \\
        EE alignment & $\text{cos}^2\EEori$ & 0.75 \\
        Grasping & $\mathbb{I}\{\text{grasp}\}$ & 1.0 \\
        \midrule
        Command rate & $60\shapingcoef\cdot\|\command - \lastcommand\|^2$ & -0.1 \\
        EE command rate & $30\shapingcoef\cdot\|\commandee - \lastcommandee\|^2$ & -0.1 \\
        Command smoothness & $30\shapingcoef\cdot\|\command - 2\lastcommandee + \lastlastcommand\|^2$ & -0.1 \\
        Collision & $\mathbb{I}\{\text{non-gripper collision}\}$ & -1.0 \\
        \bottomrule
    \end{tabular}
 }
    \label{table:reward}
    \vspace{-6mm}
\end{table}

The observation of the high-level actor includes the proprioception, encoded proprioception history, SAFE features of the handle, and $\mixest$. 
The proprioception consists of joint position, joint position target, previous command, base linear and angular velocity, EE pose, projected gravity vector, and finger distance of the gripper.
The base linear velocity is estimated from the low-level controller using proprioception history~\cite{nahrendra2023dreamwaq}.
Additionally, a vector from the EE to the handle and the heading angle of the robot are included in the observation.

\subsubsection{Reward Function}
\label{sec:reward_function}

The reward function consists of three components: opening, auxiliary, and shaping rewards.

The opening reward is designed to encourage the robot to open the target articulated object.
Therefore, it is proportional to the movement of the object, $\movement$, which is defined as the opening angle for revolute joints and the displacement for prismatic ones.
The angle and displacement are normalized, $\normmove$, using the corresponding joint limits.

The auxiliary reward guides the robot to grasp the handle in an appropriate pose.
It consists of three components: EE approach, EE alignment, and grasping reward.
The EE approach reward encourages the robot to approach its EE close to the center of the handle.
To avoid conflicts with the opening reward, it is set to the maximum value after the opening begins.
The EE alignment reward encourages the robot to align the dorsal-palmar axis of EE with the longer side of the handle, which is the proper grasping pose.
The grasping reward is given when the gripper grasps the handle.

Finally, the shaping reward is designed to prevent inappropriate motions, such as aggressively changing commands.
The command change rate and its derivative are penalized.
Note that the relative amount of shaping reward becomes smaller as the training progresses.
Thus, we increased the scaling factor of the shaping reward as the average success rate, $\shapingcoef$, rose. 
The detailed equation and scales of reward functions are summarized in Table~\ref{table:reward}.


\section{Experiments}
\label{sec:experiment_setup}
\subsection{Experiment Setup}
\label{sec:training_env}

\begin{figure}[t] 
    \vspace{3mm}
    \centering
\hspace{-9mm}
\captionsetup{font=small}
    \includegraphics[width=0.46\textwidth]{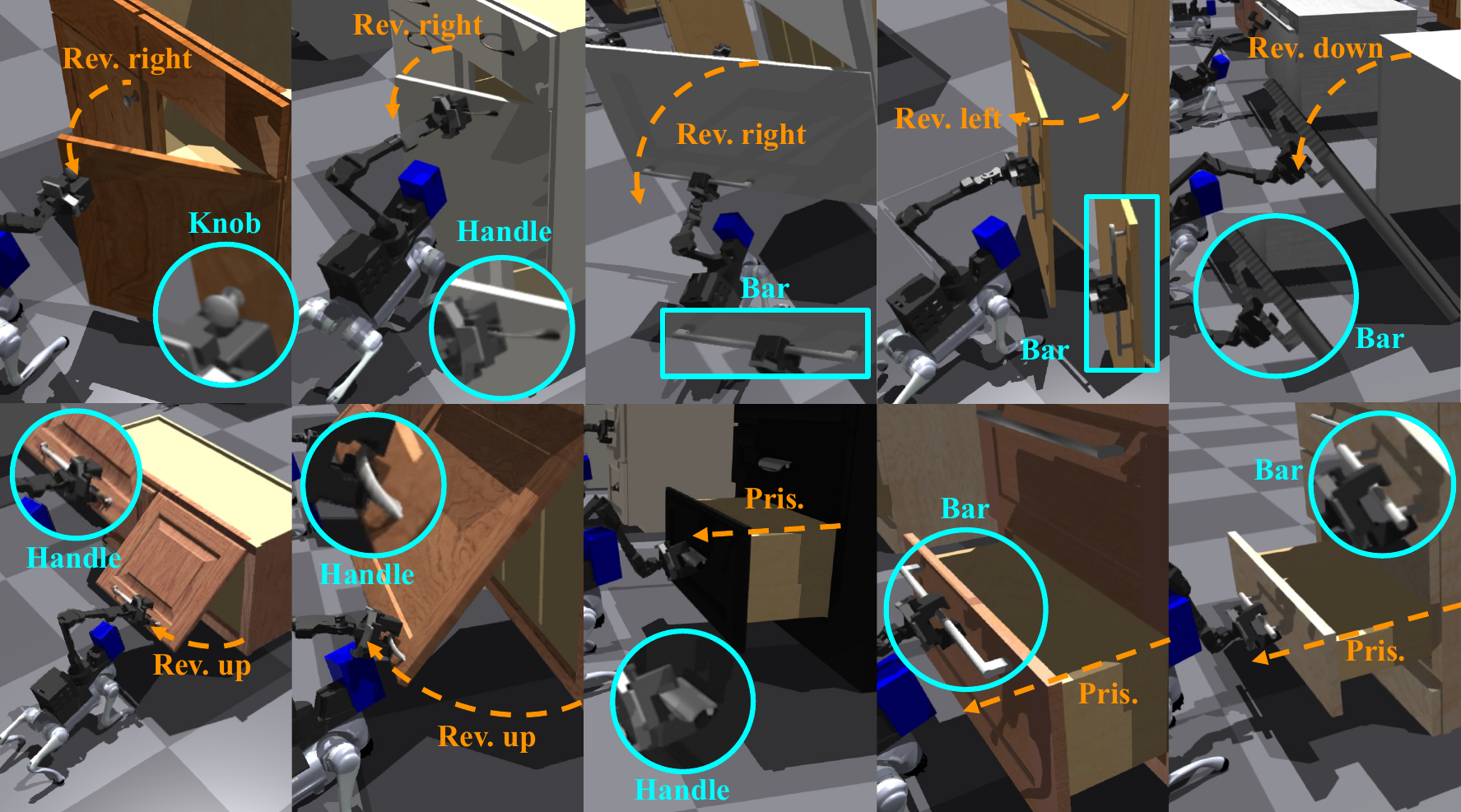}
\vspace{-4mm}
    \caption{
    Demonstration of a heterogeneous articulated object opening in the simulation.
    The objects differ in their handle shape, dimension, and opening direction.
 }
    \label{fig:demo}
\vspace{-11mm}
\end{figure}

We trained all high-level modules in the Isaac Gym~\cite{makoviychuk2021isaac}.
The training environment consisted of 41 heterogeneous articulated objects selected from the PartManip dataset~\cite{geng2023partmanip}.
The objects were selected to cover diverse combinations of articulation and handle types.
The articulation types include revolute joints with axes opened up, down, left, or right, and prismatic joints.
The handle types were heuristically classified into bar, handle, and knob depending on their shape.
Examples of each articulation and handle type are illustrated in Fig.~\ref{fig:demo}.
The relative pose between the robot and the object was randomly sampled at the beginning of each episode, while assuming that the robot is facing the object.
The initial pose of the robot was sampled within a 2 m box, centered 2 m away from the object, with a random heading angle in $[-\pi/4, \pi/4]$.
Additionally, the friction coefficient of the manipulator was set to zero, preventing the robot from scrubbing the object, as referred to Zhang \textit{et al}.~\cite{zhang2025learning}.
Instead, to compensate for the limited grasping force and friction of the finger, a small hook is attached to the tip of the gripper.

Training was performed in $2,\!048$ parallel environments, with different object assets distributed across them.
The high-level actor was trained using Proximal Policy Optimization (PPO)~\cite{schulman2017proximal}.
Similar to~\cite{rudin2022learning}, an adaptive learning rate based on the KL-divergence was used for the high-level actor and its history encoder.
Additionally, the learning rate was set to 0.0005 for ArtIEst modules, since the diversity of openable objects increases during training, and the proprioception-based estimation should update accordingly.
All training was performed on a desktop PC with an Intel Core i9-13900 CPU (up to 5.6 GHz), 48 GB RAM, and an NVIDIA RTX 4090 GPU\@.

The robot platform comprises a 12-DoF Unitree Go2~\cite{unitree_go2} legged robot and a 6-DoF ViperX 300 arm~\cite{vx300s}, which is equipped with a parallel two-finger gripper.
Additionally, two Livox Mid-360 LiDARs and an XSens MTi-300 IMU are mounted on the robot for perception.
The high-level planner, odometry, and perception algorithm were run on an ASUS NUC 13 Pro while the low-level controller was run on a separate NVIDIA Orin NX\@.
The high-level planner and low-level controller were run at 50 Hz.
We accumulated 25 steps of proprioception at 0.02s intervals for the history encoder.

The corner positions of the handle and panel cuboids were provided in the fixed world frame at the beginning of each experiment and not updated during manipulation.
Instead, their relative positions in the robot base frame were updated at 10 Hz through LiDAR odometry and provided to the high-level planner.

\subsection{Compared Methods}

We compare our method with the following baselines:
\begin{itemize}
\item \textbf{Center-based teacher~\cite{zhang2025learning}}:
This baseline was adapted from the method proposed by Zhang \textit{et al}.~\cite{zhang2025learning}. 
The door was represented using only the center position of the handle, whereas our method incorporates the handle and panel shape.
The position of the doorway is excluded, since our task does not require the robot to traverse it.
Privileged information, such as GT $\artinfo$ and joint states, was included in the observations. 
Furthermore, due to differences in task settings and robot hardware, this baseline was trained with our reward function and low-level controller.

\item \textbf{Point cloud-based policy}: 
This baseline represents the object using a high-dimensional point cloud. 
We sampled $1,\!024$ points on the frontal surface of the object.
A custom lightweight version of PointNet~\cite{qi2017pointnet} is used as a feature extractor.
We reduced the number of hidden layers of the PointNet to 2, and set the dimensions to 64 and 128 due to the large memory usage, but it still uses about three times larger GPU memory than \textbf{Ours}.
\end{itemize}
Additionally, to investigate the effectiveness of each component in our method, we conduct an ablation study with the following variants:
\begin{itemize}
\item \textbf{Ours w/o handle shape}: 
This variant excludes the handle shape feature, while still providing the center of the handle to the high-level policy.
\item \textbf{Ours w/o sampling}: 
Sampling process is ablated in this variant, thus the corner points of the enveloping cuboid are used to represent the handle and panel shape.
\item \textbf{Ours w/~mono.~est.}: 
A monolithic estimator is employed for calculating $\artinfo$ instead of separating modules as in ArtIEst. 
This monolithic estimator is similar to that of Zhang \textit{et al}.~\cite{zhang2025learning}, which jointly takes both proprioception and exteroception as inputs and estimates $\artinfo$.
\item \textbf{Ours w/o prop.~est.}: 
Without the proprioception-augmented estimator, only the exteroception-based estimator is used, similar to the existing vision-based approaches~\cite{zeng2021visual, yu2024gamma,wang2024rpmart}.
\end{itemize}

\begin{figure}[!t] 
    \vspace{2mm}
    \centering
    \hspace{-10mm}
    \vspace{-2mm}
    \captionsetup{font=small}
    \includegraphics[width=0.4 \textwidth]{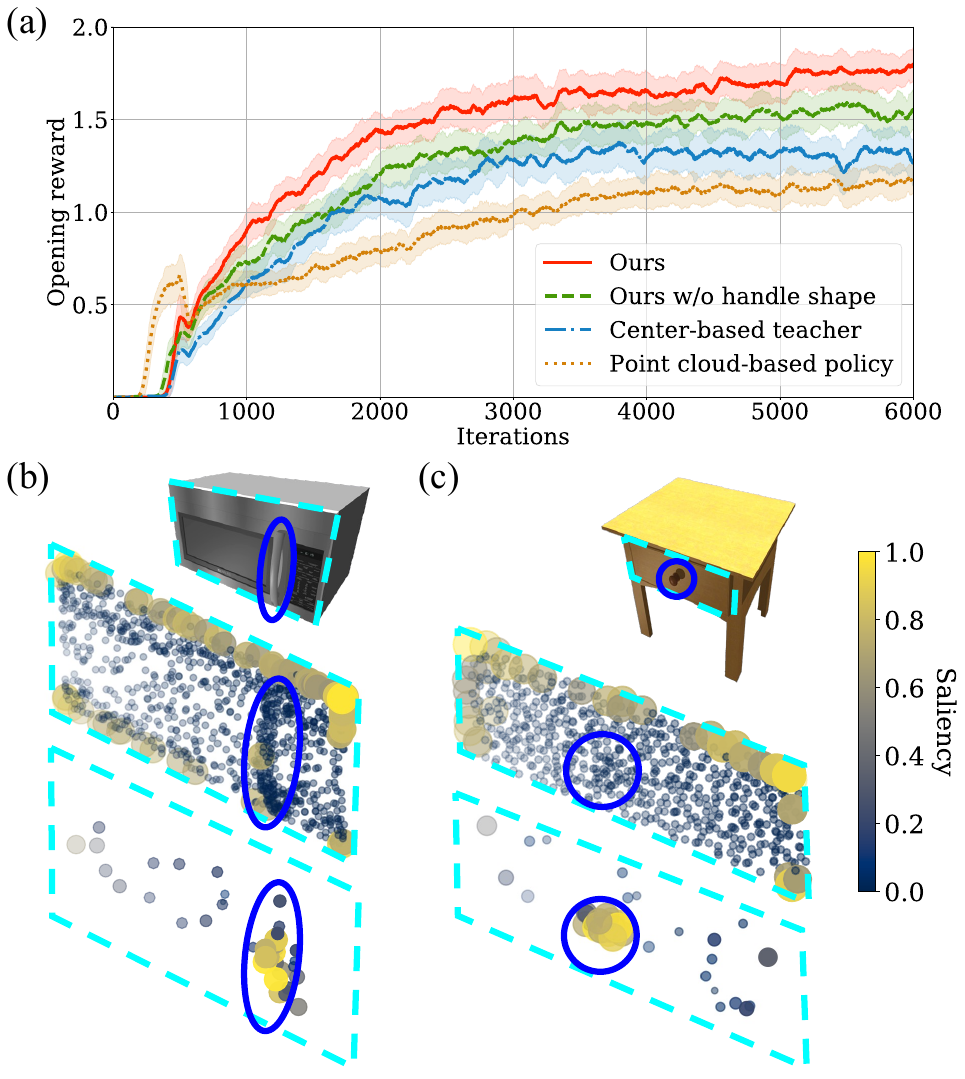}
    \caption{
(a) Learning curve of opening reward. 
\textbf{Ours} shows the highest performance compared with the baselines.
Saliency maps are shown for the object with (b)~revolute joint and (c)~prismatic joint.
Object appearances are depicted at the top line, the saliency maps from \textbf{Point cloud-based policy}, and \textbf{Ours} are shown in the middle and below each subfigure.
}
    \label{fig:learning_curve}
    \vspace{-10 mm}
\end{figure}

All the methods above were trained with the same reward functions and training environment described in Sec.~\ref{sec:reward_function} and Sec.~\ref{sec:training_env}, respectively.
For a fair comparison, exponential linear units (ELUs)~\cite{clevert2015fast} were used as the activation function in all networks.
\subsection{Efficiency of the Proposed Representation}
\begin{figure*}[t] 

\vspace{2mm}
    \centering
    \hspace{-10mm}
    \captionsetup{font=small}
    \includegraphics[width=0.885\textwidth]{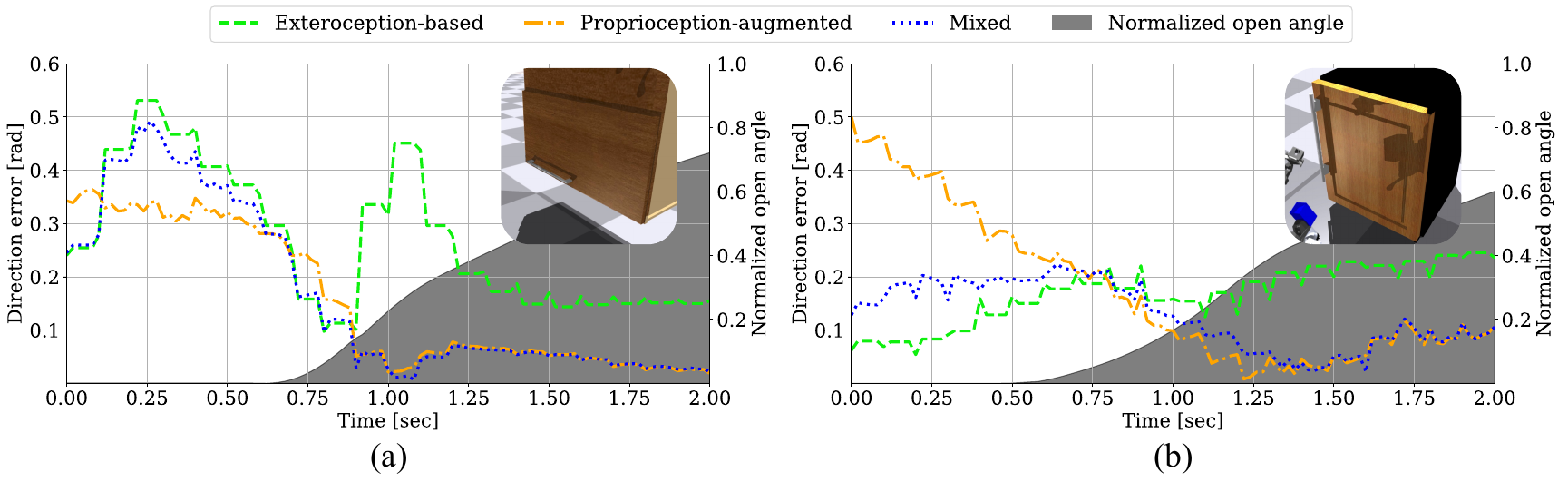}
\vspace{-2mm}
    \caption{
Estimation error transition of exteroception-based, proprioception-augmented, and mixed estimations during the manipulation.
The progress of the opening motion is indicated by the object's open angle normalized using the joint limits.
(a) The appearance of the object can be inferred to be opened in either a rightward or upward direction, while the correct direction is upward, leading to visual ambiguity in exteroception-based estimation.
(b) In contrast, no visual ambiguity exists in this case.
}
\vspace{-10 mm}
    \label{fig:belief_transition_artiest}
\end{figure*}

Fig.~\ref{fig:learning_curve} (a) shows that \textbf{Ours} achieves a higher opening reward compared with the baselines, highlighting the efficiency of the proposed representation.
Although privileged information, including $\artinfo$, is provided for the \textbf{Center-based teacher}, representing an object with handle centers appears insufficient for heterogeneous articulated objects, resulting in lower rewards.
Another difference between \textbf{Ours} and \textbf{Center-based teacher} is the inclusion of encoded proprioception history, which may help the policy to capture contact-rich dynamics between the robot and the object.
The comparison with \textbf{Ours w/o handle shape} further clarifies the role of the encoded proprioception history.
Since  \textbf{Ours w/o handle shape} retains proprioception history but excludes handle-shape features, its efficiency falls between \textbf{Ours} and \textbf{Center-based teacher}.
This suggests that both handle-shape representation and proprioception history play important roles in effective learning.

Figs.~\ref{fig:learning_curve} (b) and (c) show the saliency map of the actions of the \textbf{Point cloud-based policy} and \textbf{Ours} from the objects with (b)~revolute joint and (c)~prismatic joint at the top of each subfigure.
The saliency is obtained by computing the gradient of the action with respect to the input point cloud for \textbf{Point cloud-based policy} (middle) and sampled points for \textbf{Ours} (below).
Higher saliency indicates higher attention of the policy to a specific region in the input when making decisions.
Subsequently, the saliency map is obtained by projecting the accumulated saliency over 100 steps (2 s) onto the corresponding points.
Higher saliency values are depicted with larger points and brighter colors.
\textbf{Ours} consistently focuses on the handle shape for both objects.
However, the \textbf{Point cloud-based policy} typically focuses on the edges of the objects rather than the handle shape.
This results in slower convergence and lower performance of the \textbf{Point cloud-based policy} than that of \textbf{Ours} as shown in Fig.~\ref{fig:learning_curve} (a).
These results indicate that directly using high-dimensional observations is less efficient for learning contact-rich tasks with legged manipulators.

\subsection{Heterogeneous Articulated Object Manipulation}

We demonstrated heterogeneous articulated object manipulation with a legged manipulator in the simulation environment.
As shown in Fig.~\ref{fig:demo}, our single versatile policy successfully opens a variety of articulated objects.
The objects differ in handle shape, dimension, and articulation type.
Therefore, the robot needs to adapt its manipulation strategy to each object.
In particular, the robot has to align its gripper to grasp the handle firmly and determine the opening direction and range of motion for opening the object.
The proposed representation, encoding handle shape and opening direction, enables heterogeneous articulated object manipulation.

\subsection{Belief Gating Mechanism of ArtIEst}

Fig.~\ref{fig:belief_transition_artiest} shows the error transition of $\extest$, $\propest$, and $\mixest$ during manipulation, (a) with visual ambiguity in the articulation information and (b) without ambiguity.
The errors are measured as the angle difference between the estimated and GT articulation information.
When there is visual ambiguity, multiple candidates for $\artinfo$ are inferred, depicted as the oscillating error of $\extest$ in Fig.~\ref{fig:belief_transition_artiest}~(a).
However, errors of $\propest$ decrease as the manipulation proceeds, since proprioception during the contact can help resolve the visual ambiguity.
Therefore, the belief gate shifts the belief from $\extest$ to $\propest$, leading to a decrease in the error of $\mixest$.
In contrast, when there is no ambiguity in the exteroception-based estimation, the error of $\extest$ is low and stable, as shown in Fig.~\ref{fig:belief_transition_artiest}~(b).
However, $\propest$ shows lower error from the beginning of the contact.
Therefore, the belief transitions from $\extest$ to $\propest$.

\subsection{Modular Architecture of ArtIEst}
\begin{table}[!t]
    \centering
    \vspace{2mm}
    \captionsetup{font=small}
    \caption{
        ArtIEst evaluation.
        The estimation errors are separately measured before and during the contact.
    }
    \vspace{-1mm}
    {\scriptsize
        \begin{tabular}{lccc}
            \toprule
            \multicolumn{1}{c}{\multirow{2}{*}{\textbf{Method}}} & 
            \multicolumn{3}{c}{\textbf{Estimation direction error (rad) $\downarrow$}} \\ 
            \cmidrule(lr){2-4}
            & \textbf{Before contact} & \textbf{During contact} & \textbf{Entire episode} \\ 
            \midrule \midrule
            \textbf{Ours}                 & \textbf{0.2293} & 0.0687 &  \textbf{0.1701} \\
            Ours w/o prop.~est.  & 0.2623          & 0.2228 & 0.2482  \\ 
            Ours w/~mono.~est.   & 0.2574          & \textbf{0.0606} & 0.1827  \\
            \bottomrule
        \end{tabular}
    }
    \vspace{-6mm}
    \label{table:artiest_evaluation}
\end{table}

\begin{figure*}[!t] 
\vspace{2mm}
    \centering
    \captionsetup{font=small}
    \hspace{-9mm}
    \includegraphics[width=0.835\textwidth]{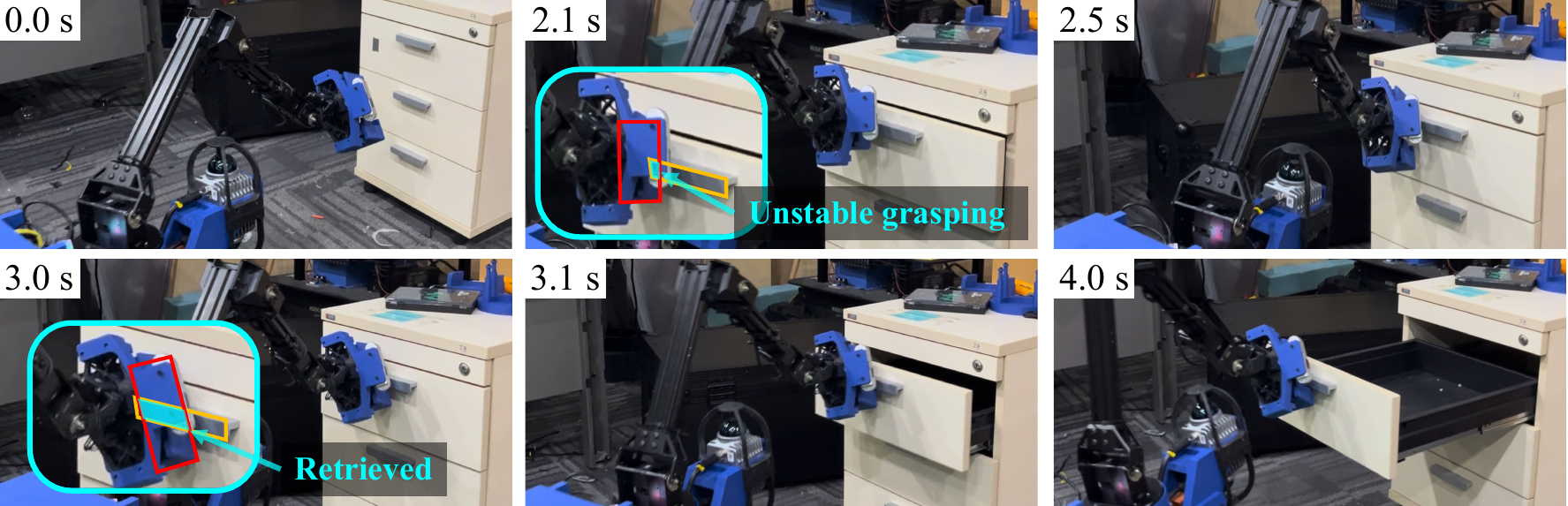}
    \vspace{-1mm}
    \caption{
    Real-world demonstration of opening a drawer with a prismatic joint and a horizontally elongated handle.
    Although the first grasp pose (2.1~s) was not stable, the robot autonomously retrieved the handle (3.0~s) and successfully opened the drawer.
}
    \vspace{-10mm}
    \label{fig:real_demo}
\end{figure*}

We evaluated the contribution of each component within ArtIEst.
The estimation errors are measured across test object sets and averaged over five episodes.
The test set was randomly selected from the dataset~\cite{geng2023partmanip}, excluding the objects used for training.
To investigate the role of excluding proprioception before contact, the estimation error was separately measured before and during the contact.

As shown in Table~\ref{table:artiest_evaluation}, \textbf{Ours} achieves the lowest estimation error for the entire episode among the methods.
Compared with \textbf{Ours w/o prop.~est.}, the consistently reduced error before and during contact demonstrates the benefit of incorporating proprioceptive information.
Additionally, \textbf{Ours w/ mono.~est.} shows higher error than \textbf{Ours} when there is no contact.
This indicates that excluding proprioceptive inputs before contact helps improve the estimation accuracy.
Although \textbf{Ours w/~mono.~est.} shows slightly lower error during contact, the difference is smaller than that of before contact, and the overall accuracy is still lower than \textbf{Ours}.

\subsection{Cross-domain Generalization of SAFE}
\begin{table}[t!]
    \centering
    \vspace{1mm}
    \captionsetup{font=small}
    \caption{Success rate across training and test sets.
    The proposed method shows the highest performance retention.}
    {\scriptsize
    \begin{tabular}{lccc}
        \toprule
        \multicolumn{1}{c}{\multirow{2}{*}{\textbf{Method}}} & 
        \multicolumn{2}{c}{\textbf{Success rate (\%)  $\uparrow$}} & 
        \multirow{2}{*}{\textbf{Test/Train~(\%) $\uparrow$}} \\ 
        \cmidrule(lr){2-3}
        & \textbf{Train set} & \textbf{Test set} &  \\
        \midrule \midrule
        \textbf{Ours} & \textbf{79.35} & \textbf{79.02} & \textbf{99.35} \\
        Ours w/o sampling & 77.56 & 70.23 & 92.92 \\ 
        \midrule
        Center-based teacher~\cite{zhang2025learning} & 62.43 & 50.60 & 81.05 \\ 
        Point cloud-based policy & 56.58 & 41.39 & 73.15 \\ 
        \bottomrule
    \end{tabular}
    }
    \vspace{-6mm}
    \label{table:safe_robustness}
\end{table}

To evaluate the enhanced cross-domain generalization resulting from the sampling process of SAFE method, we compared the success rate over the training and test sets.
An episode was considered successful if the robot opened the object at least 33\% of its maximum range of motion, which is typically a stricter criterion than that used in dataset~\cite{geng2023partmanip} (20\% for drawers and $30^{\circ}$ for doors, while doors usually can be opened more than $90^{\circ}$).
The success rate was measured over ten episodes for each object, with a maximum length of 15 s per episode.

Table~\ref{table:safe_robustness} shows that \textbf{Ours} achieves the highest success rate on both the training and test sets.
Additionally, the proposed method demonstrates robust cross-domain generalization, as evidenced by a 99.35\% \generatio~ratio.
Compared with \textbf{Ours w/o sampling}, the sampling process in SAFE improves the \generatio~ratio by 6.43\%, confirming that the sampling process contributes to enhanced generalization.
\textbf{Point cloud-based policy} shows the lowest \generatio~ratio, indicating its limited ability to generalize.
This is because directly using high-dimensional visual input without sufficient abstraction can make the policy overfit to the specific appearance details of the objects, as depicted in Figs.~\ref{fig:learning_curve} (b) and (c).

\subsection{Real-world Demonstration}

We deployed the proposed framework on a real-world legged manipulator.
The robot successfully opens a cabinet with a revolute joint and a vertical handle (Fig.~\ref{fig:title}~(a)) as well as a prismatic drawer with a horizontal handle (Fig.~\ref{fig:title}~(b)).
Although these real-world objects were not included in the training dataset, the robot successfully opened them, demonstrating the generalization capability of the proposed framework.

The sequence of opening motions for the drawer is demonstrated in Fig.~\ref{fig:real_demo}.
Although the first grasping (2.1~s) was not stable due to the misaligned gripper, the robot autonomously regrasped the handle (3.0~s) and successfully opened the drawer.
Notably, such auto-retrying behaviors are important for the robot to adapt to unexpected changes in the environment and successfully complete the task.
However, such behaviors can be difficult to achieve with model-based approaches, as they require precise modeling of contact uncertainties and adaptive replanning, which is often impractical in real-world settings.

\section{Conclusion}
\label{sec:conclusion}
In this study, we presented a framework for opening heterogeneous articulated objects with a legged manipulator.
We proposed a low-dimensional representation to extract features from the objects, enabling sample-efficient training.
The proposed framework demonstrated the ability to open diverse objects with a single policy.
The estimation error for articulation information is decreased with ArtIEst architecture, and SAFE shows improved cross-domain generalization.
Currently, the framework was demonstrated with a fixed object pose measurement given at the beginning of the episode.
For future work, we seek to integrate the framework with onboard object pose estimation.

\bibliographystyle{URL-IEEEtrans}

\bibliography{URL-bib}

\end{document}